\documentclass[pmlr]{jmlr}
\usepackage[utf8]{inputenc}
\usepackage{color}
\usepackage{times}
\usepackage{graphicx}
\usepackage{amsmath}
\usepackage{amssymb}
\usepackage{enumitem}
\usepackage{import}
\usepackage{booktabs}
\usepackage{multirow}

\usepackage{algorithm}

\DeclareMathOperator*{\argmin}{arg\,min}
\DeclareMathOperator*{\argmax}{arg\,max}

\usepackage{graphicx}
\newcommand{\VDDxSmall}{\textbf{Sim-250}}
\newcommand{\VDDxBig}{\textbf{Sim-630}}

\usepackage[noend]{algpseudocode}
\firstpageno{1}
\jmlrvolume{85}
\jmlryear{2018}
\jmlrworkshop{Machine Learning for Healthcare}

\begin{document}
\title[From expert systems to machine-learned diagnosis models]{Learning from the experts:\\ From expert systems to machine-learned diagnosis models}
       
\author{\Name{Murali Ravuri} \Email{murali@curai.com}
       \AND
       \Name{Anitha Kannan}  \Email{anitha@curai.com}
       \AND
       \Name{Geoffrey J. Tso}  \Email{geoff@curai.com}
       \AND
       \Name{Xavier Amatriain}  \Email{xavier@curai.com}
       }


\maketitle
\begin{abstract}

Expert diagnostic support systems have been extensively studied. The practical applications of these systems in real-world scenarios have been somewhat limited due to well-understood shortcomings, such as lack of extensibility. More recently, machine-learned models for medical diagnosis have gained momentum, since they can learn and generalize patterns found in very large datasets like electronic health records. These models also have shortcomings - in particular, there is no easy way to incorporate prior knowledge from existing literature or experts. In this paper, we present a method to merge both approaches by using expert systems as generative models that create simulated data on which models can be learned. We demonstrate that such a learned model not only preserves the original properties of the expert systems but also addresses some of their limitations. Furthermore, we show how this approach can also be used as the starting point to combine expert knowledge with knowledge extracted from other data sources, such as electronic health records.
\end{abstract}

\section{Introduction}
In 2015, the Institute of Medicine (IOM) called for a ``significant re--envisioning of the diagnostic process" and reported that ``nearly every person will experience a diagnostic error in their lifetime" [\cite{iom_2015}]. The medical diagnostic process, defined as ``a mapping from a patient’s data (normal and abnormal history, physical examination, and laboratory data) to a nosology of disease states" [\cite{miller_diagnostic_2016}], leads to a \emph{differential diagnosis} (DDx) consisting of a ranked list of possible diagnoses that is used to guide further assessments and possible treatments. Unsurprisingly, diagnostic error can have significant effects on the care a patient receives, and the IOM noted that these errors account for 10\% of deaths and up to 17\% of hospital adverse events. 

The thought that physicians can benefit from clinical decision support systems (CDS) specialized for diagnosis has led to substantial work in this area [\cite{miller_internist-1_1982, barnett_dxplain:_1987}]. Practical use of these systems, however, has been constrained by several factors [\cite{Miller94problems}]. A major issue is that most CDS are expert systems constructed on knowledge bases derived from medical studies. These medical studies are costly to run and may be limited in their size and their ability to generalize to the population as a whole. Maintenance of these CDS systems involves careful monitoring of existing literature, which is a laborious manual process.

The recent advent of access to digital resources such as electronic health records holds promise as a major avenue for timely access to highly granular patient-level data. Large data repositories with detailed medical information can be mined through machine learning (ML) techniques to automatically learn fine-grained diagnostic models. These machine-learned models can capture patterns at different levels of granularity, and they can be easily extended and updated, as new data becomes available. On the downside, directly incorporating  prior medical knowledge gathered through clinical research into machine learned models is difficult. Therefore, these ML models will only be as good as the data on which they are trained.

In this paper, we show that we can get the best of both worlds. More concretely, we present a novel approach to medical decision support that involves: 
\begin{enumerate}
\item Using existing expert systems as generative models to create synthetic training data;
\item Using state-of-the-art deep learning approaches to train a model on synthetic data from expert systems.
\end{enumerate}

By combining these two steps, we show that the resulting approach:
\begin{itemize}
\item is more accurate;
\item is more resilient to noise than the original expert system;
\item is more flexible and can be expanded to incorporate data from other sources.
\end{itemize}

\paragraph{Technical Significance:} This paper introduces a novel approach to combining existing traditional expert systems for medical decision support with machine-learned models. The proposed method is a way to incorporate existing prior knowledge encoded in expert systems, using a data centric approach: Expert systems serve as a data generator and the resultant synthetic data can be used as inputs for machine learning models. This technique presents a number of technical advantages such as the possibility to combine the synthetic data with other data sources such as electronic health records. 

\paragraph{Clinical Relevance:} The overall goal of the work presented in this paper is to propose a method for enhancing the performance of medical decision support systems. Our approach using machine learning models shows a significant improvement in accuracy over existing state-of-the-art expert systems, which usually require a laborious manual process to improve. If implemented in real-life scenarios, such methods could result in significantly better diagnoses and therefore better clinical outcomes.



\section{Related work}\label{sec:related}


\noindent\textbf{Expert diagnostic systems:} Expert-based systems are computer systems that emulate the decision making ability of human experts [\cite{JacksonExpertSystems}]. They generally include two components: a knowledge base that is usually created by experts, and an inference engine [\cite{ExpertSystemsProbabilities}]. Many of the early applications of such expert systems were in the medical domain, and a number of them were specifically built to support diagnosis. The first of such systems, Mycin [~\cite{mycin}], focused on infectious diseases. Successor  systems such as Internist-1 [\cite{miller_internist-1_1982}], QMR [\cite{rassinoux_modeling_1996}], and DXplain [\cite{barnett_dxplain:_1987}] broadened their focus to general internal medicine. All of these systems generated a differential diagnosis based on algorithms that utilized expert and evidence-based medical knowledge [\cite{miller_diagnostic_2016}]. See \S~\ref{sec:expert_system} for an explanation of one such expert system we use in this paper. 

A typical limitation of any expert system is the \emph{knowledge acquisition} problem [\cite{gaines_knowledge_2013}]: creating the knowledge base is a time-consuming and complex process that requires access not only to well-trained domain experts, but also to experts in modeling. This is also true for expert-based \emph{CDS}, for which adding one disease requires weeks of work from expert physicians who are also familiar with the process (see [\cite{Miller1986TheIM}]). As a result, maintaining, updating, and extending these systems can be difficult.

All of the expert systems described above were designed to support physicians in their medical decisions. As such, they were built with the assumption that the user has \emph{expert-level} medical knowledge. In recent years, simplified diagnostic systems, sometimes called symptom checkers, have also been used for \emph{user-facing} applications. These basic patient systems have poor accuracy [\cite{semigran_comparison_2016}] and are missing many of the advanced, but hard-to-use, capabilities of physician-facing CDS. 


~

\noindent\textbf{Diagnosis as an inference on expert systems:} Expert systems have been interpreted using a probabilistic formulation [\cite{ExpertSystemsProbabilities}]. In this formulation, diseases and manifestations form a bipartite probabilistic graph with the conditional probabilities and priors derived from the variables in the expert system. The goal of these formulations is to perform higher fidelity probabilistic inference on the graph. However, exact computation of the posterior distribution over the diseases, given the current manifestation of findings, is intractable due to higher-order stochastic dependencies. Much of the work in this area has focused on  approximate inference, including Markov chain Monte Carlo stochastic sampling [\cite{ShweCooper90}], variational inference [\cite{JaakolaJordan99}]  and  learning a recognition network [\cite{Morris01}] by exploiting various properties (e.g. sparsity) of the graph instantiation. In this paper, we instead pose diagnosis as a classification task by using data simulated from an expert system to train a model that is not only robust in inference, but also provides an elegant way to add data from other sources such as electronic health records.  In this way we can incrementally update the model with new diseases, or with diseases that have different manifestations different contexts (geographies, demographies, seasons, etc.).

~

\noindent\textbf{Machine learned diagnosis models:} 
With the widespread use and availability of electronic health records (EHRs), there has been a surge of research in learning diagnostic models from patient data. In [\cite{Wang_LR_RiskPrediction}], a multilinear sparse logistic regression model is used for disease onset prediction. In [\cite{rotmensch_learning_2017}], a probabilistic noisy-OR model was learned to infer a knowledge graph from electronic health records, which can be used for diagnosis, similar to probabilistic inference in expert system.
More recently, emphasis has been placed on directly learning diagnostic models, using deep neural networks, either single time shot or through time ( \emph{c.f.} [\cite{miotto_deep_2016, Ling_DeepRL_Diagnostic, Shickel_Deep_EHR, rajkomar_scalable_2018}] and references therein). 
An important distinction from our work is that these diagnostic models do not have access to any expert system or prior medical knowledge. 

\pagebreak

\section{Expert System for diagnosis}
\label{sec:expert_system}
The expert system we use in this work is the latest evolution of the original Internist-1 [\cite{miller_internist-1_1982}], and its later reincarnation as QMR [\cite{miller_qmr}]. The algorithm for producing differential diagnoses has a set of expert curated rules that operate on an underlying knowledge base of disease profiles. 

\noindent\textbf{Knowledge base:} The knowledge base consists of diseases, findings, and their relationships. A finding can be a symptom, a sign\footnote{Signs are defined as an indication of a medical condition that can be objectively observed}, a lab result, a demographic variable such as age and gender, or a relevant piece of past medical history. Each edge between a finding and a disease is parametrized by two variables: \emph{evoking strength} and \emph{frequency}. Each of these variables has an associated score, on a scale of 1 to 5, that is assigned by physician experts using their own clinical judgment and supporting medical literature.


The \emph{evoking strength} variable is a finding-disease relationship score that indicates the strength of association between its constituent finding-disease pair. While assigning evoking strength scores, physicians answered the question: ``Given this finding, how strongly should one consider this disease?" As an example, a history of alcoholism has a large evoking strength to alcoholic hepatitis, a liver disease caused by chronic alcohol ingestion. In contrast, fever, which occurs in many diseases (including alcoholic hepatitis), does not have as much evoking strength for alcoholic hepatitis.

The \emph{frequency} variable gives a complementary datum of how often patients with the given disease also have the finding. As an example, for the same alcoholic hepatitis disease, leg swelling may have low frequency, while fever, which is common amongst those suffering from alcoholic hepatitis, would have high frequency.

Each finding also has a disease-independent \emph{import} variable that represents the global importance of the finding -- \emph{i.e.}, the extent to which one is compelled to explain its presence in any patient. As an example, ``chest pain" has much higher import than a ``headache". Given a patient with these two symptoms the resulting differential diagnosis needs to consider ``chest pain" more than ``headache" due to high import of the former.

~

\noindent\textbf{Differential diagnosis algorithm:}  For the purpose of this paper, we present a succinct explanation of the algorithm for differential diagnoses from [\cite{miller_internist-1_1982}], assuming all the  findings are provided ahead of time, and the goal is to obtain the list of possible diagnoses. Interested readers can refer to  [\cite{miller_internist-1_1982}] for more in depth discussion.

The algorithm first converts the 5-scale values to non-linear weights: evoking strengths 1 to 5 is mapped to 1, 10, 20, 40 and 80 points, respectively. Similarly, frequencies and import between 1 and 5 are mapped to 1, 4, 7, 15 and 30 points.

Given a set of input findings, each disease is ranked based on a combined score with four components. The positive component weights each finding according to the evoking strength of the finding to the disease.  The negative component down-weights the score based on findings (and weighted frequencies) that are expected to occur in patients with the disease but are absent in the patient under consideration. When there are findings that are present in the patient but not explained by the disease, negative penalties are included based on a non-linear weighting of the import of the finding. If the disease corresponds to one the patient has a history of, an additional bonus award is incurred that is 20 points times the frequency number listed for the diagnosis in the disease profile.

\section{Simulating medical cases from an expert knowledge base}
\label{sec:case_simulator}
We use a medical case simulator implemented on the extended QMR knowledge base described in the previous section. An important aspect of the case simulator is that it makes a strong closed world assumption, in which the universe of diseases and findings is limited to those in the knowledge base. This closed world assumption and the structure of knowledge base lends itself to a simple but effective algorithm for simulating medical cases. The advantage of such a simulator is the ability to generate a large number of cases that can be used to train the machine-learned model described in \S~\ref{sec:model}. 

The simulation algorithm is based on [\cite{qmr-simulated-cases}], and can be described briefly as follows: The simulator first samples a disease and then constructs a set of findings in a sequence of steps. First, demographics variables are sampled based on their weighted frequency in relation to the disease in question. Then predisposing factors for the disease are sampled, again based on frequency. The rest of the findings are processed in decreasing order of their frequencies in relation to the disease - under these constraints, each finding is randomly chosen to be present or absent. If chosen to be present, then findings that are impossible to manifest are removed from consideration (\emph{e.g.} a female patient can not be both pregnant and not pregnant), and findings with high co-occurrence probabilities (\emph{e.g.} fever and chills tend to co-occur) are prioritized as candidates for inclusion, even if they have lower frequency than the next finding(s) to be considered. The simulation of the case ends when all findings in the knowledge base have been considered.

Table~\ref{tbl:cases} shows examples of realistic medical cases generated using this approach. For each disease, we provide two different cases, demonstrating the variability in their manifestations. We can see that different attributes of the same concept are explicitly called out. As an example, for the disease ``hepatitis acute viral", key attributes of vomiting that are relevant for the disease are spelled out: the vomiting needs to be recent, and it needs to look like coffee grounds. In contrast, values that require numeric understanding are bucketed into ranges: For example, for the disease ``pyrogenic shock'', arterial systolic pressure needs to be less than 90 while diastolic needs to be less than 60. 


\begin{table}[h]
\centering
\begin{tabular}{|l|}
\hline
\begin{minipage} [t]  {1.0\textwidth} 
	\vspace{.05pt}
   \noindent \textbf{Hepatitis acute viral}
   \begin{itemize}
        \item jaundice, abdomen pain exacerbation with food, abdomen pain epigastrium, hepatomegaly present, liver enlarged moderate,liver tender on palpation, abdomen pain present, joint pain mild or moderate, abdomen tenderness present
        \item anorexia, jaundice, abdomen pain epigastrium, hepatomegaly present, liver enlarged moderate, liver tender on palpation, feces light colored, hands palmar erythema, skin spider angiomas, abdomen pain acute, abdomen pain present, abdomen pain not colicky, vomiting recent, constipation, vomiting coffee grounds
   \end{itemize} 
\end{minipage} \\ \vspace*{0.1pt}\\ [3pt]  \hline  
\begin{minipage} [t] {1.0\textwidth} 
	\vspace{.05pt}
    \noindent \textbf{Arthritis acute septic} 
    \begin{itemize}
        \item joint tenderness swelling redness, joint involvement polyarticular asymmetrical, hip pain unilateral or bilateral, joint pain severe, joint range of motion decreased, knee pain unilateral or bilateral, joint effusions single or multiple, onset abrupt, fever, joint exam abnormal
        \item tachycardia, joint exam abnormal, joint tenderness swelling redness, joint pain severe, joint range of motion decreased, knee pain unilateral or bilateral, joint effusions single or multiple, joint involvement monoarticular, onset abrupt, shoulder pain left, shoulder pain right
    \end{itemize} 
\end{minipage} \\ \vspace*{0.1pt}\\[3pt]  \hline  
\begin{minipage} [t]{1.0\textwidth} 
	\vspace{.05pt}
   \noindent  \textbf{Leukemia chronic myelocytic}
    \begin{itemize}
        \item skin lesions present, lymph nodes enlarged, mouth mucosa petechiae, tourniquet test positive, gingiva hemorrhage, chest percussion diaphragm elevated unilateral, splenomegaly present
        \item hepatomegaly present, skin cutaneous nodules, rectal exam tenderness diffuse, lymph nodes enlarged, mouth mucosa petechiae, tourniquet test positive, gingiva hemorrhage, retina hemorrhages deep round, retina hemorrhage subhyaloid, retina roth spots, splenomegaly present, fever
  \end{itemize} 
\end{minipage} \\ \vspace*{0.1pt}\\[3pt]  \hline  
\begin{minipage} [t] {1.0\textwidth} 
\vspace{.05pt}
     \noindent  \textbf{Pyrogenic shock}
    \begin{itemize}
        \item coma, cyanosis of mucous membranes, tachypnea, pressure arterial systolic less than 90, skin sweating increased generalized, pressure arterial diastolic less than 60, myalgia, fever, tachycardia, rigors
        \item cyanosis of mucous membranes, fever,  pulse arterial thready, coma, rigors, pressure arterial systolic 90 to 110, pressure arterial diastolic less than 60, skin sweating increased generalized
        \vspace{.05pt} 
     \end{itemize} 
    \end{minipage} \\[3pt] \hline  
\end{tabular}
\caption{Examples of medical cases simulated for various diseases using the approach in \S~\ref{sec:case_simulator}.}
\label{tbl:cases}
\vspace{-10pt}
\end{table}





\section{Machine learning models}\label{sec:model}

The case simulator described in the previous section provides a large training set of medical cases: Each case $c$ is associated with a set of findings, $\mathbf{s}_c$ such that $s_i \in \mathbf{F}$ manifested in a simulated patient. During training, we also have the corresponding diagnosis for each case $y_c \in \mathbf{D}$ where $\mathbf{F} = \{f_1, ... f_N\}$ and  $\mathbf{D} = \{d_1, ... d_K\}$  are the universe of all findings and diseases, respectively.

Following the standard supervised learning paradigm, the goal of the machine learned models we describe next is to learn parameters $\theta$ that maximize the likelihood of the correct diagnosis:
\begin{equation}
\theta^{*} = \argmax_{\theta} \sum_{(\mathbf{s}, y)} \log p(y | \mathbf{s}; \theta)
\end{equation}

We would like to further show that such a model can replicate, and at times improve, the diagnostic performance of expert systems, from which the data has been simulated. \\


\noindent{\textbf{Finding representation}} All findings in $\mathbf{F}$ are tokenized into a set of unigram tokens. Let $\mathbf{Z}$ be the universe of all unigram tokens in $\mathbf{F}$, the set of findings, $\mathbf{s}$, associated with a medical case is also tokenized with $ z(\mathbf{s}) \subset \mathbf{Z}$ being the union of these tokens.  The collection of pairs $\{\mathbf{z}(\mathbf{s}^{(t)}), y^{(t)}\}_{t=1}^{T}$ serves as the training data of $T$ medical cases for the two models we describe next.

\subsection{Multiclass logistic regression}
As mentioned in \S~\ref{sec:related}, logistic regression is a simple linear model that has been used for diagnosis in prior works. For this reason, we choose this model as a baseline machine learned model for the multiclass classifier. 

Multiclass logistic regression  learns a linear mapping between $\mathbf{z}$ and label $y$, with the mapping function defined by:
\begin{equation}
P(y = k|\mathbf{s},W, \mathbf{b}) = \frac{\exp (b_k+\mathbf{z(\mathbf{s})}. \mathbf{w}_k)}{1 + \sum_{j} \exp(b_j+\mathbf{z(\mathbf{s}} . \mathbf{w}_j))}
\end{equation}
where $\mathbf{w}_j$ is a vector of weights and $b_k$ is the bias for the $k^{th}$ class. 
$\mathbf{z(\mathbf{s})}$ is a sparse $|\mathbf{Z}|$ dimensional binary vector such that it has 1 in dimensions that correspond to an element in  $\mathbf{z}(\mathbf{s})$.

Given a training set, $\{\mathbf{z}^{(t)}, y^{(t)}\}_{t=1}^{T}$, the model is trained using stochastic gradient descent by minimizing the negative log-likelihood (including an L2 regularization term):
\begin{equation}
W^*, b^* = \argmin_{W, b} - \sum_t\log P(y^{(t)}|\mathbf{s}^{(t)},W, \mathbf{b}) + \lambda \sum_{j} |\mathbf{ w}_{j}|^2,
\end{equation}
where $\lambda = 0.01$ was chosen empirically using performance on validation set.

\subsection{Deep neural network model}
Our second approach is a Deep Neural Network (DNN) model that is also learned directly from the tokens in the findings. We choose a deep learning approach, given the recent successes of this family of models (see \S~\ref{sec:related}) mainly because of their ability to capture higher order relationships in the input that are meaningful for the task at hand. In our setting, interactions between various components of a patient's medical profile, such as symptom-symptom or medical history-symptom symptoms and medical history, are highly pertinent. Instead of hand-crafting these dependencies, with deep models, we can have the model automatically capture these relationships through layers of interaction between variables.

Convolutional neural networks (CNN) have recently shown to be effective for text processing tasks [\cite{Collobert_text_processing}]. Here, we use a convolutional architecture adapted to our problem space. The model takes as input $\mathbf{z}(\mathbf{s})$ and predicts the diagnosis.  The overall architecture is presented in Table \ref{fig:model} and includes:
\begin{itemize}
\item An embedding layer that converts each $z_i \in \mathbf{z}(\mathbf{s})$ into embedding 
vector through a lookup table and then concatenates them.  
\item Three temporal convolutional layers, each followed by a ReLU activation. The first layer transforms the embeddings into feature maps. The subsequent layers learn higher order representations of the features.
\item The outputs from the last layer are flattened and interleaved with dropout and fully connected layers. 
\item The final layer is a fully connected layer, with softmax activation, to 
output the probability of each diagnosis.
\end{itemize}
\begin{table}
\centering
\begin{tabular}{r c c l}
\toprule
\textbf{Layer} &  \textbf{(Filters, size)} &  \textbf{Output}  \\
 \midrule
Embedding &  & $200 \times 50$\\
Temporal Convolutional $|$ ReLU & (256, 3) &  $ 200 \times 256$  \\
Temporal Convolutional $|$ ReLU & (128, 3) &  $ 200 \times 128$  \\
Temporal Convolutional $|$ ReLU & (64, 3) &   $ 200 \times 64$  \\
Flatten & &  12800 \\
Dropout (0.5) $|$ Fully connected $|$ ReLU & $12800 \rightarrow 180$  & 180  \\
Dropout (0.5) $|$ Fully connected $|$ Softmax  & $180 \rightarrow K$& K (\# of classes)  \\
\bottomrule
\end{tabular}
\caption{Convolutional model used in our experiments. Input is 200 dimensional word sequence. }
\label{fig:model}
\vspace{-10pt}
\end{table}
\noindent {\textbf{Training details}: During training, the objective is to maximize the
log probability of the correct diagnosis. The model is trained with minibatches of size 1024 using Nesterov accelerated stochastic gradient descent  with initial learning rate of 0.1 and momentum 0.9. All parameters of the model are updated at each step of the optimization. The parameters, including embeddings, are initialized randomly.
Class imbalances are handled using random sampling with replacement.


\section{Experiments}

\subsection{Experimental setup}

\noindent\textbf{Dataset:}  
For all of the experiments, we used the case simulator described in \S~\ref{sec:case_simulator}.
The experiments were performed on the following base datasets:
\begin{itemize}
\item \VDDxSmall:  This a dataset of 400,000 cases simulated from a reduced knowledge base with 250 prevalent diseases and 3343 findings. The breakdown of finding types in the knowledge base is: past medical history (481), symptoms (211), signs (961) and labs (1691).


\item \VDDxBig: This is a dataset of 1,000,000 cases simulated from the entire knowledge base with 630 diseases and 4600 findings. The breakdown of manifestation types include: past medical history (528), symptoms (222), signs (1141) and labs (3664).
\end{itemize}
Along with \VDDxSmall and \VDDxBig, we also created corresponding noisy counterparts for testing model efficacy in the presence of noise. Details of this are provided in \S~\ref{sec:noise_robustness}.

We used approximately 8:1:1 as train-test-validation split, while ensuring that every disease had at least 100 cases in the test set. We also ensured that no case in the test set had an exact replica in the training set. This leads to class imbalance in the training set that was handled through sampling with replacement.


~

\noindent\textbf{Metrics:} We use top-1 and top-3 accuracy as the metrics for comparing different models. In particular, top-k accuracy (a.k.a. recall@k) over a test set T is given by:
\begin{equation}
\textrm{top~k accuracy} = \frac{\sum_{t=1}^{T} \sum_{j=1}^{j=K}[y^{(t)} = \hat{y}^{(t)}[j]]}{T},
\end{equation}
where $[a=b]$ is the Iverson notation that evaluates to one only if a=b or else to zero. $\hat{y}^{(t)}[j]$ is the $j^{th}$ top class predicted from a model when evaluating test case $t$.
\\
~

\noindent\textbf{Baseline: Differential diagnosis as probabilistic inference:}
Following rich literature on interpreting frequency strength between the disease and finding as a class-conditional probability of an observation (c.f. [\cite{JaakolaJordan99}]),  we constructed a probabilistic generative model of findings in which 
findings are independent given the diseases: $p(\mathbf{f} | d =k ) = \prod_{j} p(f_j| d=k)$. Probabilities are set by normalizing the non-linear weighting scheme (\S.~\ref{sec:model}) to be valid probabilities. We performed posterior inference on this model using mean field variational approximation so that:
\begin{equation}
p(d=k|\mathbf{f}) \propto p(\mathbf{f} | d =k ) p(d=k) \label{eq:bayes_rule}
\end{equation}
While being a simple baseline, it is also congruent to the simulator used in creating the dataset, where only the frequencies are used for data generation. Note though that this model ignores the relationships and higher-order dependencies implicit in posterior inference.

\subsection{Accuracy}
\label{sub:accuracy}
The goal of this initial set of experiments is to establish that models that are learned on the simulated data from the expert knowledge base can obtain comparable performance to the original expert system that was designed to operate on the knowledge base. 


We trained the ML models in two settings: with and without labs. The ``without labs" setting is akin to most patient presentations for a new illness where the physician is less likely to have access to lab work to aid in their initial differential diagnosis. The ``with labs" setting is more representative of the information available to a physician further along the diagnostic process. ``Labs" have so much information about the disease that they can serve as an indicator of the underlying cause on their own.  We report results on the \VDDxSmall~and \VDDxBig~datasets both for test sets that include and exclude the lab results.

Table~\ref{tab:noise_free_history_sign_system} presents our main results.  Key takeaways include:
\begin{itemize}
\item  DNN (with labs)  improves top-1 accuracy almost 10\% for the smaller dataset and more than 5\% in the larger data set when compared to the expert system.  When compared to the probabilistic inference baseline, the gains are over 5\% in both cases.
\item  A somewhat surprising aspect is that  DNN (with labs) shows degraded performance on the test set without labs. This can be explained by the fact that labs results are extremely discriminative between diseases and thus have overarching importance. 
\item Not surprisingly, DNN (without labs) and LR (without labs) both perform poorly on the test set with labs, as test and training data distribution are quite different. However, they perform significantly better than any other model in the test set without data.
\end{itemize}

For simplicity, the rest of the paper will use the \VDDxSmall~(without labs) dataset both for training and testing. The choice of not including labs is to understand the impact of signals that physicians have access at the time of first contact with the patient. 

\begin{table}
\centering
\begin{tabular}{c c c c c c}
\toprule
    \textbf{Dataset} & \textbf{Approach} & \textbf{@1 w/ labs} & \textbf{@3 w/ labs} & \textbf{@1 w/o labs} & \textbf{@3 w/o labs} \\
     \midrule
    \multirow{3}{*}\VDDxSmall & Expert system & 0.90& 0.98& 0.71 & 0.88\\
     & Probabilistic inf.                          & 0.94 & 0.98 & 0.82 & 0.94\\
     & LR (w/ labs)                & 0.95 & 0.99 & 0.40 & 0.56\\
     & LR (w/o labs)                & 0.65 & 0.83 & 0.73 & 0.89\\
     & DNN (w/ labs)                 & \textbf{0.99} & \textbf{0.998} & 0.58 & 0.71\\
     & DNN (w/o labs)                 & 0.46 & 0.60 & \textbf{0.92} & \textbf{0.98}\\
    \midrule
    \multirow{3}{*}\VDDxBig & Expert system & 0.93 & 0.96 & 0.59 & 0.81\\
    & Probabilistic inf.                           & 0.93 & 0.97 & 0.75 & 0.88\\
    & LR (w/ labs)                  & 0.81 & 0.93 & 0.25 & 0.38\\
    & LR (w/o labs)                & 0.35 & 0.55 & 0.44 & 0.63\\
    & DNN (w/ labs)                    & \textbf{0.98} & \textbf{0.998}& 0.53 & 0.66\\
    & DNN (w/o labs)                 & 0.43 & 0.56& \textbf{0.88} & \textbf{0.96}\\
     \midrule
\bottomrule
\end{tabular}
\caption{Top 1 and Accuracy comparison between the different methods; Both training and test data are noise-free. Results shown here for 
the two datasets setting when the observations are mix of history, signs, symptoms and labs. We include training and testing configurations that include and exclude lab results. See Tbl.~\ref{tab:noise_free_250} for a more detailed breakdown
based on the types of input observations to the system}
\label{tab:noise_free_history_sign_system}
\vspace{-10pt}
\end{table}

Next, and as an extension of the previous idea of removing lab results, we consider what would happen to the different approaches if we lacked a subset of the manifestation types during \emph{test} time. To evaluate the resilience of these methods, we considered a testing setup in which only a subset of manifestations types (signs, history, symptoms) are present at test time. Table~\ref{tab:noise_free_250} shows the results for various combinations.  We note that for all models, the most useful data is included in the signs. For any  combination that includes signs, our deep learning model beats both the expert and the probabilistic inference model. However, interestingly, for those test set combinations without signs, the probabilistic inference performs better. A possible explanation for this is that probabilistic inference directly operates on the findings; therefore it is less affected when signs are absent. In contrast, learned models, such as LR and DNN, have captured the importance of signs for discriminating between diseases. In the same vein, the expert model will penalize lack of findings that are expected to occur in patients but which are absent in the testing, as described in \S~\ref{sec:expert_system}.

\begin{table}
\centering
\begin{tabular}{ r  c c c c @{}}
\toprule
  \multirow{2}{*}{\textbf{Test time inputs to model}} &  \multicolumn{3}{c}{\textbf{Approach}}\\
     \cmidrule(lr){2-5} &  \textbf{Expert} & \textbf{Probabilistic} & \textbf{LR} & \textbf{DNN}  \\
 \midrule
symptom &0.33 (0.56) &\textbf{0.43 (0.66)} & 0.15 (0.26) &0.32 (0.47)  \\
history &0.36 (0.54)&\textbf{0.48 (0.69)} &0.12 (0.21)&0.38 (0.53) \\
history and symptom &0.48 (0.67)&\textbf{0.62 (0.79)} &0.46 (0.43)&0.54 (0.69)  \\
sign &0.58 (0.79)&0.67 (0.85)& 0.30 (0.66)& \textbf{0.72 (0.88)}  \\
history and sign &0.65 (0.84) &0.77 (0.91) &0.59 (0.76)&\textbf{0.86 (0.95)}  \\
sign and symptom &0.62 (0.82)&0.71 (0.88) &0.59 (0.79)&\textbf{0.82 (0.95)}  \\
history, sign and symptom &0.71 (0.88)&0.82 (0.94)&0.73 (0.89)&\textbf{0.92 (0.98)}  \\
    \midrule
\bottomrule
\end{tabular}
\caption{Top 1 (and top 3) accuracy for various observation configurations for the \VDDxSmall~dataset. For LR and DNN, training was done using (history, sign and symptom) configuration. Note, the probabilistic method does not involve any model training.}
\label{tab:noise_free_250}
\vspace{-10pt}
\end{table}



\begin{figure}[t!]
\centering
\includegraphics[width=6in]{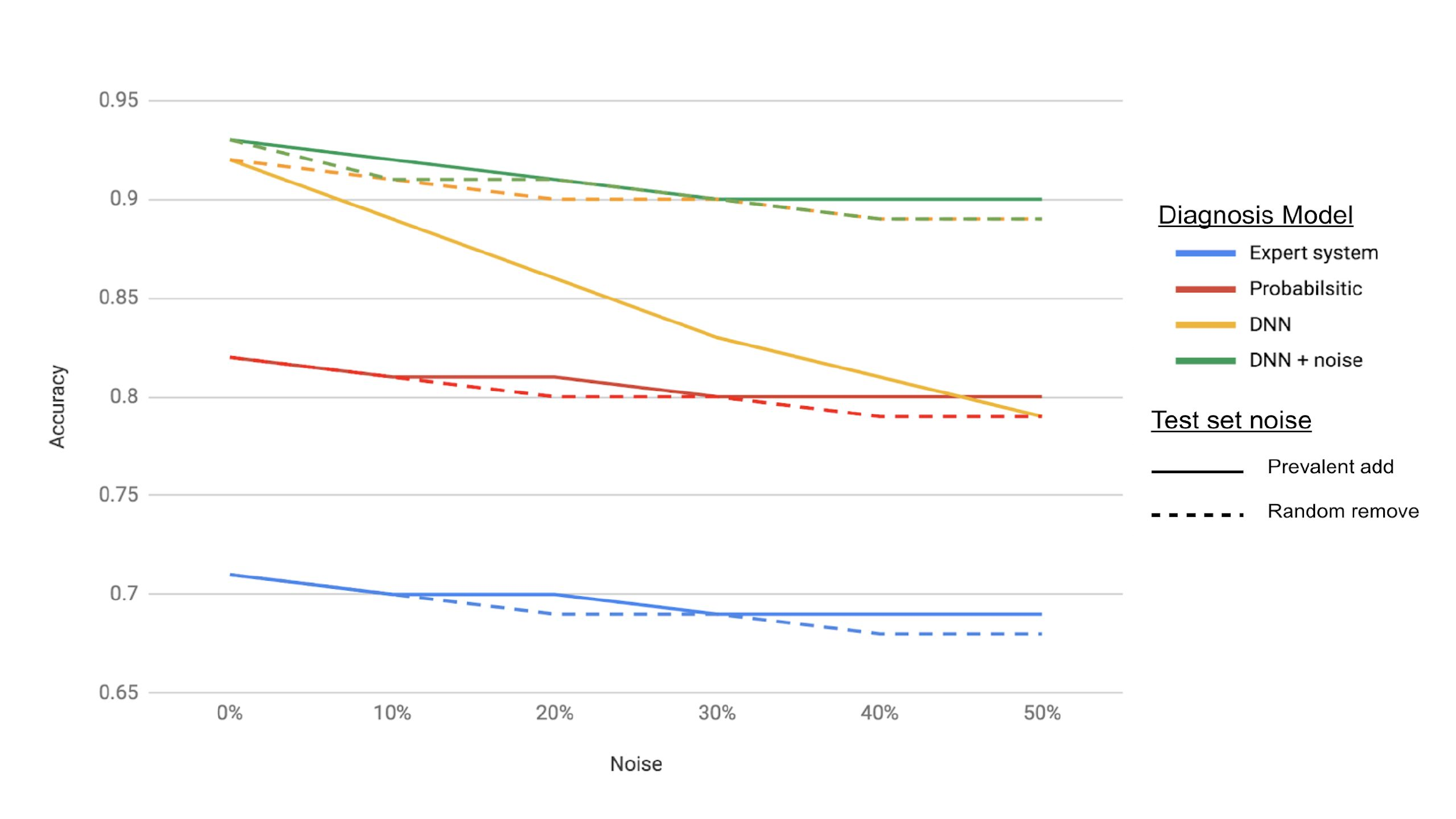}
\caption{Results on robustness of algorithms to noise (best seen in color). Color of the line corresponds to a model while, line style is reflective of type of noise (prevalent add or random remove) added.  Results shown for recall @1. We observe similar performance gaps for recall @3. We also omit results for logistic regression model for simplicity. See text for details. The 0\% noise is replicated across all noise types and copied from Tbl.~\ref{tab:noise_free_history_sign_system} for comparison.}
\label{tab:noise_robustness}
\end{figure}

\subsection{Robustness to noise}
\label{sec:noise_robustness}

In a real life situation, the input to the model may be noisy, either in the form of additional or missing findings. We would like our CDS to be as robust as possible to such noise. Because of this, we investigated the robustness of the different algorithms under varying levels of noise.


In order to mimic real-life situations as much as possible, we corrupted the original data set by (1) adding prevalent findings, and (2) removing findings at random. We did not remove generally prevalent findings since they are less likely to be missing from a disease. We did not add findings at random either since random additions can completely change the diagnosis. As an example, consider a patient that has sore throat and runny nose with diagnosis of common cold. Adding a random finding that the person ``can not breathe", changes the likely diagnosis to a severe throat infection\footnote{For rigor, we experimented with this, and found that indeed all methods including expert system had huge drop in performance when adding random findings. Given that we can not disentangle between incorrect ground truth labels and bad algorithmic performance, we do not report results.}. For each of these settings, we added or removed one to five findings from each case to generate a noisy dataset. We then combined it with the original noise-free dataset by random sampling and selecting different quantities of noisy cases in the range from 10 to 50\%. In particular, x\% noisy cases implies that x\% of the cases in the dataset have been corrupted; These x\% noisy cases are generated such that there is equal proportion of one to five findings randomly added or removed.

We hypothesized that we could train an even more robust machine learning model by injecting noise during training. In order to test this, we trained the ``DNN + noise" model, which uses the same model architecture as ``DNN", but is trained by injecting two prevalent findings at random to each case during training\footnote{We tried various noise injection configurations during training and have not noticed significant differences in performance.}. For both models, only  history, signs and symptoms are used as observations. For simplicity, we do not report results for the LR model since the LR model performed poorly when compared to DNN even in noise free situations (see \ref{sub:accuracy}).

Figure~\ref{tab:noise_robustness} compares the performance of different models under the various settings of noise levels. We make multiple observations: 
\begin{itemize}
\item ``DNN" outperforms both expert systems and the probabilistic model. With  ``DNN + noise", the performance difference is significantly higher, showcasing the robustness of a trainable model.  In addition, the lowest performing setting of random five findings addition has only 6\% drop (86.23\% from 92\% in the noise-free case) in performance compared to the noise-free setting, both at train and test times. 
\item Expert systems are resilient to prevalent add and random remove of findings. As described in \S~\ref{sec:expert_system}, every finding not pertinent to a disease gets a negative score based on import. By design, on average, prevalent findings typically have low evoking strength and import and thus is unlikely to change diagnosis. Similarly, in the light of observed findings, missing a few at random, is unlikely to change diagnosis. 
\item Probabilistic inference baseline is also resilient to noise, albeit having lower performance than a learned model. This can be explained by the fact the class conditional probabilities of this inference method mimic the expert system, and given that  non-pertinent findings only contribute based on their frequency, they do not affect as much. In addition, due to  multiplicative scoring, the performance has certain level of robustness especially when incorporating a small number of random findings.
\item While ``DNN" drops in performance as we increase the amount of noise, we find that with ``DNN+noise", we can gain back the robustness of the learned model by adding only two prevalent findings at random during model training.
\end{itemize}

\subsection{Flexibility of the ML approach}
In this experiment, we show that machine learned models allow us to quickly and flexibly add new diseases into the system as long as there are sufficient number of cases diagnosed. To illustrate this, we used data from electronic health records (EHR). Extracting data from EHRs for diagnosis is a well-known practice both in literature and in practice (see [\cite{LiuEHRResearch}]) for an overview of many of those uses, for example). In our case, we used data from the publicly available MIMIC-III dataset (see [\cite{mimic3}]). 

~

\noindent{\textbf{Data processing from MIMIC-III}: MIMIC-III provides access to de-identified clinical notes of 55,000 patients admitted to intensive care units. We restricted to clinical notes written at the time of discharge. A discharge summary consists of textual description of symptoms, physical findings, complications the patient went through, and ends with discharge diagnosis. For this experiment, we focused on the sections corresponding to initial symptoms and the discharge diagnosis. We used an in-house entity extraction system to map textual descriptions to medical concepts corresponding to findings from the symptoms section of the record, and diseases from the discharge diagnosis section. Each of these concepts has a UMLS concept unique identifier [\cite{bodenreider_umls_2004}]. There can be multiple discharge diagnoses since patients in the intensive care unit (ICU) often have complex disease processes and may have experienced additional complications. To reduce the noise, we considered only diseases that have at least 1000 cases, and also removed diseases with low tf-idf (Example: sepsis, a commonly occurring condition among patients admitted to hospital for prolonged stays). We further processed the disease list for each case and made sure to assign only one disease through the following rule of thumb: the disease needs to be in the candidate set, and other co-occurring diseases are not prevalent in MIMIC-III.  This assignment of medical case is a noisy process since it ignores co-morbidities and also complications that patient may experience by being in the ICU and resultant diagnosis. This is further compounded by small sample size. We leave the problem of extracting reliable diagnostic labels from discharge summary as future work.
%
%

~

\noindent{\textbf{New diseases added from MIMIC-III:} The above data processing  resulted in seven new diseases to add to our machine learned model. Out of these, two diseases already existed in \VDDxSmall, albeit with slightly different names. We processed those to ensure they were added as new examples of cases of the existing diseases instead of a new disease.  We kept 10\% of samples from each disease for testing and then combined the remaining with \VDDxSmall~dataset simulated cases as the new data set on which we trained ``DNN'', from scratch. One can also envision extensions to the model that would allow adding new diseases, and we leave that as future work.

~

\noindent{\textbf{Results:} The first row in table \ref{tab:accuracy_on_mimic3_testset} presents the performance results when the trained model was evaluated only on the cases for those 7 diseases from the MIMIC-III test set\footnote{Note that overall accuracy on a combined test set remains almost unchanged since it is dominated by the test cases generated synthetically}. We can see a significant reduction, especially in top-1 accuracy compared to  Table \ref{tab:noise_robustness}. 

~

\noindent{\textbf{Analysis:} To understand if the drop in performance was due to noisy mappings between findings and diseases in MIMIC-III (see data description above), we did the following experiment: we considered two diseases: prematurity (of neonate), and subdural hematoma. Prematurity was chosen since it is not defined in our knowledge base of adult diseases. Subdural hematoma is present in the knowledge base, but has more specific electronic and clinical phenotype since MIMIC-III is based on ICU patients. The disease profile defined in our knowledge base is generalized to all patients diagnosed with a subdural hematoma, including those that do not require hospital care. In addition, we found that these MIMIC-III cases often were less likely to be complicated by other co-existing medical conditions or complications.

We re-trained models combining cases from these two diseases with \VDDxSmall~dataset. On this newly trained model with these two additional diseases, top-1 accuracy on just MIMIC-III test cases was 0.99 while the accuracy was unaffected for the combined or simulated-only test cases (results are reported in the second row of table~\ref{tab:accuracy_on_mimic3_testset}. This indicates that the performance drop can be attributed to the noisy data set. 

Through this experiment, we have shown as a proof of concept that it is possible to generalize to new diseases and their findings; We are also able to add new disease profiles specific to a particular phenotypic variation. We leave the modeling and optimization of the hybrid expert systems with (potentially noisy) cases from electronic health records to future work. 

\begin{table}
\centering
\begin{tabular}{c c c c}
\toprule
    \textbf{Train set} &\textbf{Test set} & \textbf{Metric} & \textbf{value} \\
     \midrule
     \multirow{2}{*}{\VDDxSmall  + 7 MIMIC-III diseases} & \multirow{2}{*}{MIMIC-III test cases 2 diseases} & top-1 accuracy & 0.63\\
     & & top-3 accuracy                          & 0.90\\
        \midrule
    \multirow{2}{*}{\VDDxSmall  + 2 MIMIC-III diseases} & \multirow{2}{*}{MIMIC-III test cases from 2 diseases} & top-1 accuracy & 0.99\\
     & & top-3 accuracy                          & 0.99\\
\bottomrule
\end{tabular}
\caption{Top-K Accuracy on EHR-only test cases when training model on a combination of expert simulated data and EHR data.}
\label{tab:accuracy_on_mimic3_testset}
\vspace{-10pt}
\end{table}
\section{Conclusion and Future work}

Medical diagnosis has been a prototypical application of AI that has attracted attention from researchers focused on classical expert systems as well as the latest machine learning approaches. Both approaches have their own pros and cons. In this paper, we introduce a methodology to combine these approaches that not only preserves the properties of the original expert system, but also improves accuracy and flexibility. One of the highlights of our proposed approach is that it enables combining data generated by an expert knowledge base with data coming from real-world medical records.

We believe this represents a novel first step in a new and promising direction. However, there are several limitations and opportunities to explore and improve on: We have used a single expert system and knowledge base, and two representative machine-learned models. In future work, we would like to extend this approach to incorporate multiple expert systems and ensembles of machine learned models. 

Similarly, we have illustrated the flexibility of our approach by using an EHR dataset with a smaller set of diseases, and its specificity to the ICU reduced its complexity in terms of multiple phenotypic manifestations of each disease and common co-morbid medical conditions. However, in order to fully validate this finding, we would like to do larger scale experiments with more diverse datasets. 
%

Finally, we would like to acknowledge that while the accuracies reported (recall @1 and @3) can be interpreted as a reasonable proxy of the effectiveness of the different diagnosis support approaches, we would like to perform more exhaustive studies that should include expert validation by physicians as well as statistical significance analysis.
\pagebreak

\bibliography{curai}

\end{document}